\documentclass{article}

\usepackage{times}
\usepackage{graphicx} 
\usepackage{subfigure}

\usepackage{natbib}

\usepackage{amsmath}
\usepackage{amsfonts}
\usepackage{url}

\usepackage[bookmarks=false]{hyperref}

\newcommand{\prednet}{g}
\newcommand{\servo}{f}
\newcommand{\img}{\mathbf{I}}
\newcommand{\grasp}{\mathbf{v}}
\newcommand{\pose}{\mathbf{p}}
\newcommand{\success}{\ell}

\usepackage[accepted]{icml2016}

\newcommand{\nwc}{\newcommand}
\nwc{\as}{\textrm{a.s.}}
\nwc{\defas}{:=}

\nwc{\dist}{\ \sim\ }
\nwc{\distiid}{\stackrel{\mathrm{iid}}{\sim}}

\icmltitlerunning{Learning Hand-Eye Coordination for Robotic Grasping with Deep Learning and Large-Scale Data Collection}

\begin{document}

\twocolumn[
\icmltitle{Learning Hand-Eye Coordination for Robotic Grasping with Deep Learning and Large-Scale Data Collection}

\icmlauthor{Sergey Levine}{slevine@google.com}
\icmlauthor{Peter Pastor}{peterpastor@google.com}
\icmlauthor{Alex Krizhevsky}{akrizhevsky@google.com}
\icmlauthor{Deirdre Quillen}{dequillen@google.com}
\icmladdress{Google}

\icmlkeywords{robotics, deep learning, neural networks}

\vskip 0.3in
]

\begin{abstract}
We describe a learning-based approach to hand-eye coordination for robotic grasping from monocular images. To learn hand-eye coordination for grasping, we trained a large convolutional neural network to predict the probability that task-space motion of the gripper will result in successful grasps, using only monocular camera images and independently of camera calibration or the current robot pose. This requires the network to observe the spatial relationship between the gripper and objects in the scene, thus learning hand-eye coordination. We then use this network to servo the gripper in real time to achieve successful grasps. To train our network, we collected over 800,000 grasp attempts over the course of two months, using between 6 and 14 robotic manipulators at any given time, with differences in camera placement and hardware. Our experimental evaluation demonstrates that our method achieves effective real-time control, can successfully grasp novel objects, and corrects mistakes by continuous servoing.
\end{abstract}

\section{Introduction}
\label{sec:intro}

When humans and animals engage in object manipulation behaviors, the interaction inherently involves a fast feedback loop between perception and action. Even complex manipulation tasks, such as extracting a single object from a cluttered bin, can be performed with hardly any advance planning, relying instead on feedback from touch and vision. In contrast, robotic manipulation often (though not always) relies more heavily on advance planning and analysis, with relatively simple feedback, such as trajectory following, to ensure stability during execution \cite{srinivasa2012}. Part of the reason for this is that incorporating complex sensory inputs such as vision directly into a feedback controller is exceedingly challenging. Techniques such as visual servoing \cite{handbook_of_robotics} perform continuous feedback on visual features, but typically require the features to be specified by hand, and both open loop perception and feedback (e.g. via visual servoing) requires manual or automatic calibration to determine the precise geometric relationship between the camera and the robot's end-effector.

\begin{figure}[t]
\includegraphics[width=0.99\columnwidth]{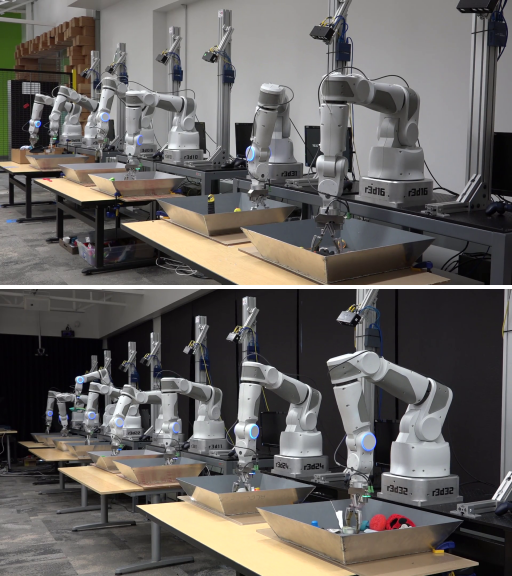}
\caption{Our large-scale data collection setup, consisting of 14 robotic manipulators. We collected over 800,000 grasp attempts to train the CNN grasp prediction model. \label{fig:teaser}}
\vspace{-0.2in}
\end{figure}

In this paper, we propose a learning-based approach to hand-eye coordination, which we demonstrate on a robotic grasping task. Our approach is data-driven and goal-centric: our method learns to servo a robotic gripper to poses that are likely to produce successful grasps, with end-to-end training directly from image pixels to task-space gripper motion. By continuously recomputing the most promising motor commands, our method continuously integrates sensory cues from the environment, allowing it to react to perturbations and adjust the grasp to maximize the probability of success. Furthermore, the motor commands are issued in the frame of the robot, which is not known to the model at test time. This means that the model does not require the camera to be precisely calibrated with respect to the end-effector, but instead uses visual cues to determine the spatial relationship between the gripper and graspable objects in the scene.

Our method consists of two components: a grasp success predictor, which uses a deep convolutional neural network (CNN) to determine how likely a given motion is to produce a successful grasp, and a continuous servoing mechanism that uses the CNN to continuously update the robot's motor commands. By continuously choosing the best predicted path to a successful grasp, the servoing mechanism provides the robot with fast feedback to perturbations and object motion, as well as robustness to inaccurate actuation.

The grasp prediction CNN was trained using a dataset of over 800,000 grasp attempts, collected using a cluster of similar (but not identical) robotic manipulators, shown in Figure~\ref{fig:teaser}, over the course of several months.
Although the hardware parameters of each robot were initially identical, each unit experienced different wear and tear over the course of data collection, interacted with different objects, and used a slightly different camera pose relative to the robot base. These differences provided a diverse dataset for learning continuous hand-eye coordination for grasping.


The main contributions of this work are a method for learning continuous visual servoing for robotic grasping from monocular cameras, a novel convolutional neural network architecture for learning to predict the outcome of a grasp attempt, and a large-scale data collection framework for robotic grasps. Our experimental evaluation demonstrates that our convolutional neural network grasping controller achieves a high success rate when grasping in clutter on a wide range of objects, including objects that are large, small, hard, soft, deformable, and translucent. Supplemental videos of our grasping system show that the robot employs continuous feedback to constantly adjust its grasp, accounting for motion of the objects and inaccurate actuation commands. We also compare our approach to open-loop variants to demonstrate the importance of continuous feedback, as well as a hand-engineering grasping baseline that uses manual hand-to-eye calibration and depth sensing. Our method achieves the highest success rates in our experiments. Our dataset is available here: {\footnotesize\url{https://sites.google.com/site/brainrobotdata/home}}

\section{Related Work}
\label{sec:related}

Robotic grasping is one of the most widely explored areas of manipulation. While a complete survey of grasping is outside the scope of this work, we refer the reader to standard surveys on the subject for a more complete treatment \cite{bohg2014}. Broadly, grasping methods can be categorized as geometrically driven and data-driven. Geometric methods analyze the shape of a target object and plan a suitable grasp pose, based on criteria such as force closure \cite{weisz2012} or caging \cite{rodriguez2012}. These methods typically need to understand the geometry of the scene, using depth or stereo sensors and matching of previously scanned models to observations \cite{goldfeder2009}.

Data-driven methods take a variety of different forms, including human-supervised methods that predict grasp configurations \cite{herzog2014,lenz2015} and methods that predict finger placement from geometric criteria computed offline \cite{goldfeder2009grasp_database}. Both types of data-driven grasp selection have recently incorporated deep learning \cite{kappler2015,lenz2015,redmon2015}. Feedback has been incorporated into grasping primarily as a way to achieve the desired forces for force closure and other dynamic grasping criteria \cite{hudson2012}, as well as in the form of standard servoing mechanisms, including visual servoing (described below) to servo the gripper to a pre-planned grasp pose \cite{kragic2002}. The method proposed in this work is entirely data-driven, and does not rely on any human annotation either at training or test time, in contrast to prior methods based on grasp points. Furthermore, our approach continuously adjusts the motor commands to maximize grasp success, providing continuous feedback. Comparatively little prior work has addressed direct visual feedback for grasping, most of which requires manually designed features to track the end effector \cite{vahrenkamp2008,hebert2012}.


Our approach is most closely related to recent work on self-supervised learning of grasp poses by \citet{lg-sss-15}. This prior work proposed to learn a network to predict the optimal grasp orientation for a given image patch, trained with self-supervised data collected using a heuristic grasping system based on object proposals. In contrast to this prior work, our approach achieves continuous hand-eye coordination by observing the gripper and choosing the best motor command to move the gripper toward a successful grasp, rather than making open-loop predictions. Furthermore, our approach does not require proposals or crops of image patches and, most importantly, does not require calibration between the robot and the camera, since the closed-loop servoing mechanism can compensate for offsets due to differences in camera pose by continuously adjusting the motor commands. We trained our method using over 800,000 grasp attempts on a very large variety of objects, which is more than an order of magnitude larger than prior methods based on direct self-supervision \cite{lg-sss-15} and more than double the dataset size of prior methods based on synthetic grasps from 3D scans \cite{kappler2015}. 

In order to collect our grasp dataset, we parallelized data collection across up to 14 separate robots. Aside from the work of \citet{lg-sss-15}, prior large-scale grasp data collection efforts have focused on collecting datasets of object scans. For example, Dex-Net used a dataset of 10,000 3D models, combined with a learning framework to acquire force closure grasps \cite{gmp-dex-16}, while the work of \citet{ot-aaib-15} proposed autonomously collecting object scans using a Baxter robot. \citet{ot-aaib-15} also proposed parallelizing data collection across multiple robots. More broadly, the ability of robotic systems to learn more quickly by pooling their collective experience has been proposed in a number of prior works, and has been referred to as collective robot learning and an instance of cloud robotics \cite{ikkhi-prrbr-00,k-cehr-10,kmckg-cbrgg-13,kpag-srcra-15}.

Another related area to our method is visual servoing, which addresses moving a camera or end-effector to a desired pose using visual feedback \cite{kragic2002}. In contrast to our approach, visual servoing methods are typically concerned with reaching a pose relative to objects in the scene, and often (though not always) rely on manually designed or specified features for feedback control \cite{ecr-navsr-92,whb-reecu-96,vahrenkamp2008,hebert2012,mkd-vbcqp-14}. Photometric visual servoing uses a target image rather than features \cite{Caron13a}, and several visual servoing methods have been proposed that do not directly require prior calibration between the robot and camera \cite{ya-auvs-94,jfn-eeuvs-97,kragic2002}. To the best of our knowledge, no prior learning-based method has been proposed that uses visual servoing to directly move into a pose that maximizes the probability of success on a given task (such as grasping).

In order to predict the optimal motor commands to maximize grasp success, we use convolutional neural networks (CNNs) trained on grasp success prediction. Although the technology behind CNNs has been known for decades \cite{lecun1995}, they have achieved remarkable success in recent years on a wide range of challenging computer vision benchmarks \cite{krizhevsky2012}, becoming the de facto standard for computer vision systems. However, applications of CNNs to robotic control problems has been less prevalent, compared to applications to passive perception tasks such as object recognition \cite{krizhevsky2012,wohlhart2015}, localization \cite{girshick2014rcnn}, and segmentation \cite{chen2014}. Several works have proposed to use CNNs for deep reinforcement learning applications, including playing video games \cite{mnih2015}, executing simple task-space motions for visual servoing \cite{lampe2013}, controlling simple simulated robotic systems \cite{wsbr-etc-15,lhph-ccdrl-16}, and performing a variety of robotic manipulation tasks \cite{levine2015}. Many of these applications have been in simple or synthetic domains, and all of them have focused on relatively constrained environments with small datasets.


\section{Overview}
\label{sec:overview}

Our approach to learning hand-eye coordination for grasping consists of two parts. The first part is a prediction network $\prednet(\img_t, \grasp_t)$ that accepts visual input $\img_t$ and a task-space motion command $\grasp_t$, and outputs the predicted probability that executing the command $\grasp_t$ will produce a successful grasp. The second part is a servoing function $\servo(\img_t)$ that uses the prediction network to continuously control the robot to servo the gripper to a success grasp. We describe each of these components below: Section~\ref{sec:prediction} formally defines the task solved by the prediction network and describes the network architecture, Section~\ref{sec:servo} describes how the servoing function can use the prediction network to perform continuous control.

By breaking up the hand-eye coordination system into components, we can train the CNN grasp predictor using a standard supervised learning objective, and design the servoing mechanism to utilize this predictor to optimize grasp performance. The resulting method can be interpreted as a type of reinforcement learning, and we discuss this interpretation, together with the underlying assumptions, in Section~\ref{sec:rl}.

In order to train our prediction network, we collected over 800,000 grasp attempts using a set of similar (but not identical) robotic manipulators, shown in Figure~\ref{fig:teaser}. We discuss the details of our hardware setup in Section~\ref{sec:hardware}, and discuss the data collection process in Section~\ref{sec:objects}. To ensure generalization of the learned prediction network, the specific parameters of each robot varied in terms of the camera pose relative to the robot, providing independence to camera calibration. Furthermore, uneven wear and tear on each robot resulted in differences in the shape of the gripper fingers. Although accurately predicting optimal motion vectors in open-loop is not possible with this degree of variation, as demonstrated in our experiments, our continuous servoing method can correct mistakes by observing the outcomes of its past actions, achieving a high success rate even without knowledge of the precise camera calibration.

\section{Grasping with Convolutional Networks and Continuous Servoing}
\label{sec:grasping}

In this section, we discuss each component of our approach, including a description of the neural network architecture and the servoing mechanism, and conclude with an interpretation of the method as a form of reinforcement learning, including the corresponding assumptions on the structure of the decision problem.

\subsection{Grasp Success Prediction with Convolutional Neural Networks}
\label{sec:prediction}

\begin{figure}
\setlength{\unitlength}{1.00\columnwidth}
\begin{picture}(1.0,0.5) \linethickness{0.5pt}
	\put(0,0){\includegraphics[width=0.49\columnwidth]{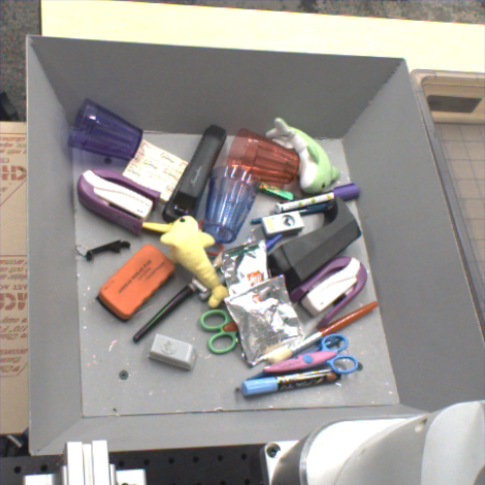}}
	\put(0.5,0){\includegraphics[width=0.49\columnwidth]{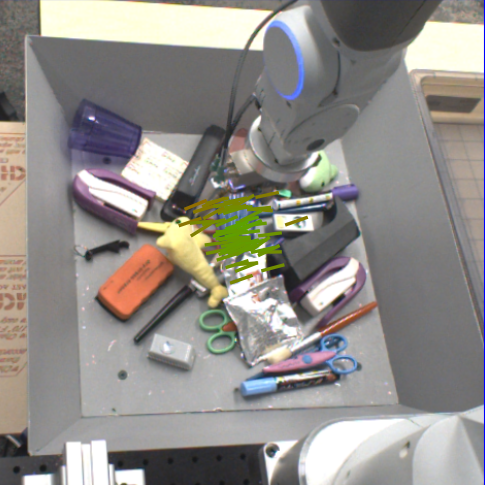}}
   \put(0.02,0.43){\large{ $\img_0$}}
   \put(0.52,0.43){\large{ $\img_t$}}
\end{picture}
	\caption{Example input image pair provided to the network, overlaid with lines to indicate sampled target grasp positions. Colors indicate their probabilities of success: green is $1.0$ and red is $0.0$. The grasp positions are projected onto the image using a known calibration only for visualization. The network does not receive the projections of these poses onto the image, only offsets from the current gripper position in the frame of the robot. \label{fig:input}}
	\vspace{-0.1in}
\end{figure}

The grasp prediction network $\prednet(\img_t, \grasp_t)$ is trained to predict whether a given task-space motion $\grasp_t$ will result in a successful grasp, based on the current camera observation $\img_t$. In order to make accurate predictions, $\prednet(\img_t, \grasp_t)$ must be able to parse the current camera image, locate the gripper, and determine whether moving the gripper according to $\grasp_t$ will put it in a position where closing the fingers will pick up an object. This is a complex spatial reasoning task that requires not only the ability to parse the geometry of the scene from monocular images, but also the ability to interpret material properties and spatial relationships between objects, which strongly affect the success of a given grasp. A pair of example input images for the network is shown in Figure~\ref{fig:input}, overlaid with lines colored accordingly to the inferred grasp success probabilities. Importantly, the movement vectors provided to the network are not transformed into the frame of the camera, which means that the method does not require hand-to-eye camera calibration. However, this also means that the network must itself infer the outcome of a task-space motor command by determining the orientation and position of the robot and gripper.

\begin{figure}
\setlength{\unitlength}{1.00\columnwidth}
\begin{picture}(1.0,0.75) \linethickness{0.5pt}
	\put(0,0){\includegraphics[width=0.99\columnwidth]{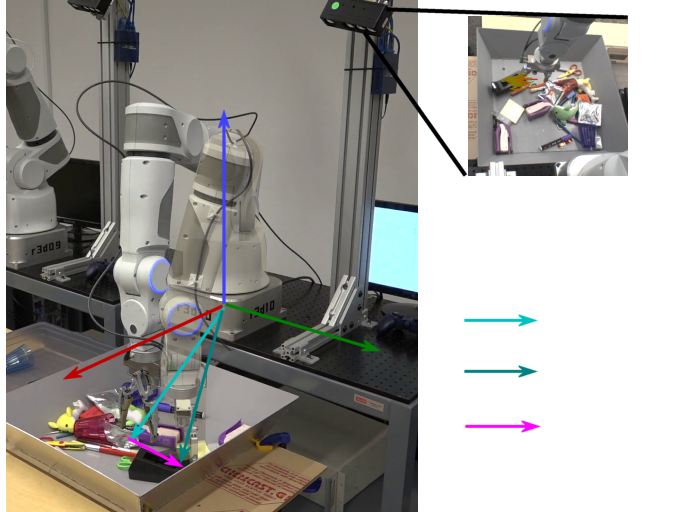}}
   \put(0.77,0.43){\Large{ $\img_t^i$}}
   \put(0.11,0.07){\Large{ $\pose_t^i$}}
   \put(0.24,0.03){\Large{ $\pose_T^i$}}
   \put(0.79,0.27){\Large{ $\pose_t^i$}}
   \put(0.79,0.19){\Large{ $\pose_T^i$}}
   \put(0.79,0.11){\Large{ $\grasp_t^i$}}
\end{picture}
	\caption{Diagram of the grasp sample setup. Each grasp $i$ consists of $T$ time steps, with each time step corresponding to an image $\img_t^i$ and pose $\pose_t^i$. The final dataset contains samples $(\img_t^i, \pose_T^i - \pose_t^i, \success_i)$ that consist of the image, a vector from the current pose to the final pose, and the grasp success label. \label{fig:grasp_diagram}}
	\vspace{-0.1in}
\end{figure}
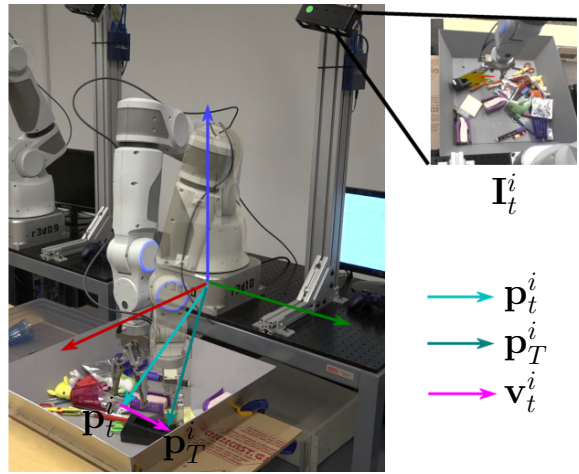

Data for training the CNN grasp predictor is obtained by attempting grasps using real physical robots. Each grasp consists of $T$ time steps. At each time step, the robot records the current image $\img_t^i$ and the current pose $\pose_t^i$, and then chooses a direction along which to move the gripper. At the final time step $T$, the robot closes the gripper and evaluates the success of the grasp (as described in Appendix~\ref{app:success}), producing a label $\success_i$. Each grasp attempt results in $T$ training samples, given by $(\img_t^i, \pose_T^i - \pose_t^i, \success_i)$. That is, each sample includes the image observed at that time step, the vector from the current pose to the one that is eventually reached, and the success of the entire grasp. This process is illustrated in Figure~\ref{fig:grasp_diagram}. This procedure trains the network to predict whether moving a gripper along a given vector and then grasping will produce a successful grasp. Note that this differs from the standard reinforcement-learning setting, where the prediction is based on the current state and motor command, which in this case is given by $\pose_{t+1} - \pose_t$. We discuss the interpretation of this approach in the context of reinforcement learning in Section~\ref{sec:rl}.

\begin{figure*}
	\includegraphics[width=2.1\columnwidth]{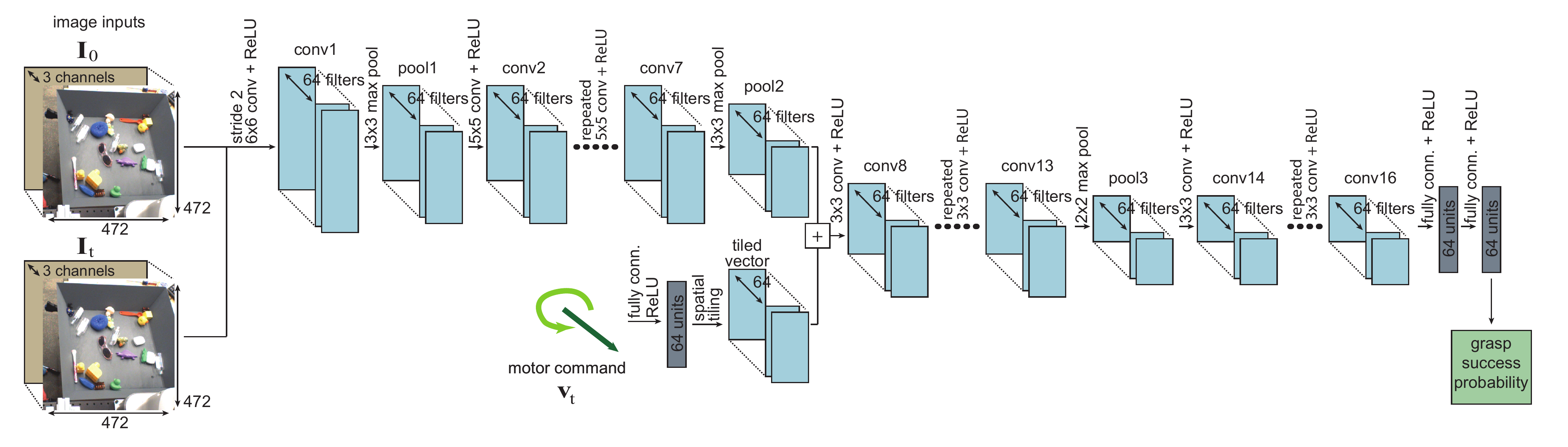}
\caption{The architecture of our CNN grasp predictor. The input image $\img_t$, as well as the pregrasp image $\img_0$, are fed into a $6 \times 6$ convolution with stride 2, followed by $3 \times 3$ max-pooling and 6 $5 \times 5$ convolutions. This is followed by a $3 \times 3$ max-pooling layer. The motor command $\grasp_t$ is processed by one fully connected layer, which is then pointwise added to each point in the response map of pool2 by tiling the output over the special dimensions. The result is then processed by 6 $3 \times 3$ convolutions, $2 \times 2$ max-pooling, 3 more $3 \times 3$ convolutions, and two fully connected layers with 64 units, after which the network outputs the probability of a successful grasp through a sigmoid. Each convolution is followed by batch normalization. \label{fig:arch}}
\vspace{-0.1in}
\end{figure*}

The architecture of our grasp prediction CNN is shown in Figure~\ref{fig:arch}. The network takes the current image $\img_t$ as input, as well as an additional image $\img_0$ that is recorded before the grasp begins, and does not contain the gripper. This additional image provides an unoccluded view of the scene. The two input images are concatenated and processed by 5 convolutional layers with batch normalization \cite{is-bnad-15}, following by max pooling. After the $5^\text{th}$ layer, we provide the vector $\grasp_t$ as input to the network. The vector is represented by 5 values: a 3D translation vector, and a sine-cosine encoding of the change in orientation of the gripper about the vertical axis.\footnote{In this work, we only consider vertical pinch grasps, though extensions to other grasp parameterizations would be straightforward.} To provide this vector to the convolutional network, we pass it through one fully connected layer and replicate it over the spatial dimensions of the response map after layer 5, concatenating it with the output of the pooling layer. After this concatenation, further convolution and pooling operations are applied, as described in Figure~\ref{fig:arch}, followed by a set of small fully connected layers that output the probability of grasp success, trained with a cross-entropy loss to match $\success_i$, causing the network to output $p(\success_i = 1)$. The input matches are $512 \times 512$ pixels, and we randomly crop the images to a $472\times 472$ region during training to provide for translation invariance.

Once trained the network $\prednet(\img_t, \grasp_t)$ can predict the probability of success of a given motor command, independently of the exact camera pose. In the next section, we discuss how this grasp success predictor can be used to continuous servo the gripper to a graspable object.

\subsection{Continuous Servoing}
\label{sec:servo}

In this section, we describe the servoing mechanism $\servo(\img_t)$ that uses the grasp prediction network to choose the motor commands for the robot that will maximize the probability of a success grasp. The most basic operation for the servoing mechanism is to perform inference in the grasp predictor, in order to determine the motor command $\grasp_t$ given an image $\img_t$. The simplest way of doing this is to randomly sample a set of candidate motor commands $\grasp_t$ and then evaluate $\prednet(\img_t, \grasp_t)$, taking the command with the highest probability of success. However, we can obtain better results by running a small optimization on $\grasp_t$, which we perform using the cross-entropy method (CEM) \cite{rk-cem-04}. CEM is a simple derivative-free optimization algorithm that samples a batch of $N$ values at each iteration, fits a Gaussian distribution to $M < N$ of these samples, and then samples a new batch of $N$ from this Gaussian. We use $N = 64$ and $M = 6$ in our implementation, and perform three iterations of CEM to determine the best available command $\grasp_t^\star$ and thus evaluate $\servo(\img_t)$. New motor commands are issued as soon as the CEM optimization completes, and the controller runs at around 2 to 5 Hz.

One appealing property of this sampling-based approach is that we can easily impose constraints on the types of grasps that are sampled. This can be used, for example, to incorporate user commands that require the robot to grasp in a particular location, keep the robot from grasping outside of the workspace, and obey joint limits. It also allows the servoing mechanism to control the height of the gripper during each move. It is often desirable to raise the gripper above the objects in the scene to reposition it to a new location, for example when the objects move (due to contacts) or if errors due to lack of camera calibration produce motions that do not position the gripper in a favorable configuration for grasping.

We can use the predicted grasp success $p(\success = 1)$ produced by the network to inform a heuristic for raising and lowering the gripper, as well as to choose when to stop moving and attempt a grasp. We use two heuristics in particular: first, we close the gripper whenever the network predicts that $(\img_t, \emptyset)$, where $\emptyset$ corresponds to no motion, will succeed with a probability that is at least $90\%$ of the best inferred motion $\grasp_t^\star$. The rationale behind this is to stop the grasp early if closing the gripper is nearly as likely to produce a successful grasp as moving it. The second heuristic is to raise the gripper off the table when $(\img_t, \emptyset)$ has a probability of success that is less than $50\%$ of $\grasp_t^\star$. The rationale behind this choice is that, if closing the gripper now is substantially worse than moving it, the gripper is most likely not positioned in a good configuration, and a large motion will be required. Therefore, raising the gripper off the table minimizes the chance of hitting other objects that are in the way. While these heuristics are somewhat ad-hoc, we found that they were effective for successfully grasping a wide range of objects in highly cluttered situations, as discussed in Section~\ref{sec:experiments}. Pseudocode for the servoing mechanism $\servo(\img_t)$ is presented in Algorithm~\ref{alg:servo}. Further details on the servoing mechanism are presented in Appendix~\ref{app:servo}.

\begin{algorithm}[h]
{\small
	\caption{Servoing mechanism $\servo(\img_t)$}
	\label{alg:servo}
	\begin{algorithmic}[1]
		\STATE Given current image $\img_t$ and network $\prednet$.
		\STATE Infer $\grasp_t^\star$ using $\prednet$ and CEM.
		\STATE Evaluate $p = \prednet(\img_t, \emptyset) / \prednet(\img_t, \grasp_t^\star)$.
		\IF{$p > 0.9$}
			\STATE Output $\emptyset$, close gripper.
		\ELSIF{$p \leq  0.5$}
			\STATE Modify $\grasp_t^\star$ to raise gripper height and execute $\grasp_t^\star$.
		\ELSE
			\STATE Execute $\grasp_t^\star$.
		\ENDIF
	\end{algorithmic}
}
\end{algorithm}

\subsection{Interpretation as Reinforcement Learning}
\label{sec:rl}

One interesting conceptual question raised by our approach is the relationship between training the grasp prediction network and reinforcement learning. In the case where $T=2$, and only one decision is made by the servoing mechanism, the grasp network can be regarded as approximating the Q-function for the policy defined by the servoing mechanism $\servo(\img_t)$ and a reward function that is $1$ when the grasp succeeds and $0$ otherwise. Repeatedly deploying the latest grasp network $\prednet(\img_t,\grasp_t)$, collecting additional data, and refitting $\prednet(\img_t,\grasp_t)$ can then be regarded as fitted Q iteration \cite{asm-fqi-08}. However, what happens when $T > 2$? In that case, fitted Q iteration would correspond to learning to predict the final probability of success from tuples of the form $(\img_t, \pose_{t+1}-\pose_t)$, which is substantially harder, since $\pose_{t+1}-\pose_t$ doesn't tell us where the gripper will end up at the end, before closing (which is $\pose_T$).

Using $\pose_T - \pose_t$ as the action representation in fitted Q iteration therefore implies an additional assumption on the form of the dynamics. The assumption is that the actions induce a transitive relation between states: that is, that moving from $\pose_1$ to $\pose_2$ and then to $\pose_3$ is equivalent to moving from $\pose_1$ to $\pose_3$ directly. This assumption does not always hold in the case of grasping, since an intermediate motion might move objects in the scene, but it is a reasonable approximation that we found works quite well in practice. The major advantage of this approximation is that fitting the Q function reduces to a prediction problem, and avoids the usual instabilities associated with Q iteration, since the previous Q function does not appear in the regression. An interesting and promising direction for future work is to combine our approach with more standard reinforcement learning formulations that do consider the effects of intermediate actions. This could enable the robot, for example, to perform nonprehensile manipulations to intentionally reorient and reposition objects prior to grasping.

\section{Large-Scale Data Collection}
\label{sec:data}

In order to collect training data to train the prediction network $\prednet(\img_t, \grasp_t)$, we used between 6 and 14 robots at any given time. An illustration of our data collection setup is shown in Figure~\ref{fig:teaser}. This section describes the robots used in our data collection process, as well as the data collection procedure. The dataset is available here: {\footnotesize\url{https://sites.google.com/site/brainrobotdata/home}}

\subsection{Hardware Setup}
\label{sec:hardware}

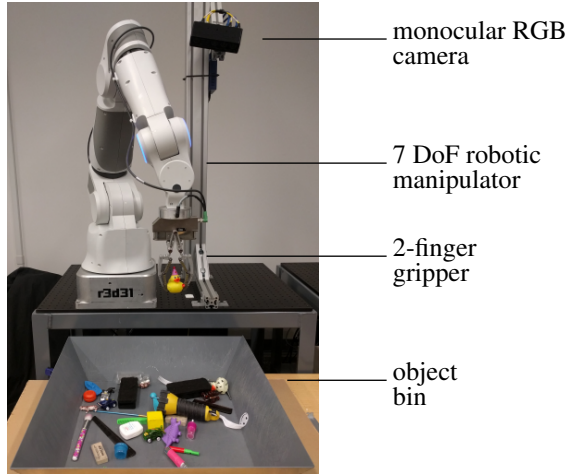
\begin{figure}
\setlength{\unitlength}{1.00\columnwidth}
\begin{picture}(1.0,0.76) \linethickness{0.5pt}
\put(0.05,0){\includegraphics[width=0.5\columnwidth]{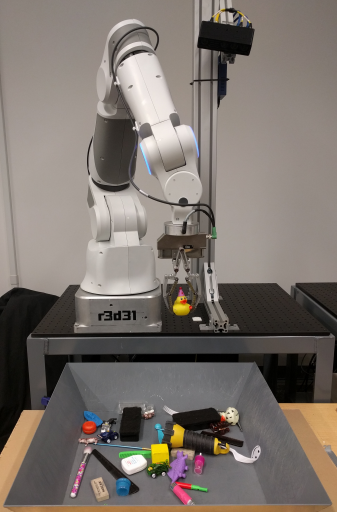}}
\put(0.46,0.7){\line(1,0){0.2}}
\put(0.37,0.5){\line(1,0){0.29}}
\put(0.37,0.35){\line(1,0){0.29}}
\put(0.50,0.15){\line(1,0){0.16}}
\put(0.67,0.7){monocular RGB}
\put(0.67,0.66){camera}
\put(0.67,0.5){7 DoF robotic}
\put(0.67,0.46){manipulator}
\put(0.67,0.35){2-finger}
\put(0.67,0.31){gripper}
\put(0.67,0.15){object}
\put(0.67,0.11){bin}
\end{picture}
	\caption{Diagram of a single robotic manipulator used in our data collection process. Each unit consisted of a 7 degree of freedom arm with a 2-finger gripper, and a camera mounted over the shoulder of the robot. The camera recorded monocular RGB and depth images, though only the monocular RGB images were used for grasp success prediction. \label{fig:singlerobot}}
	\vspace{-0.1in}
\end{figure}

\begin{figure*}
        \begin{center}
        \includegraphics[width=2.0\columnwidth]{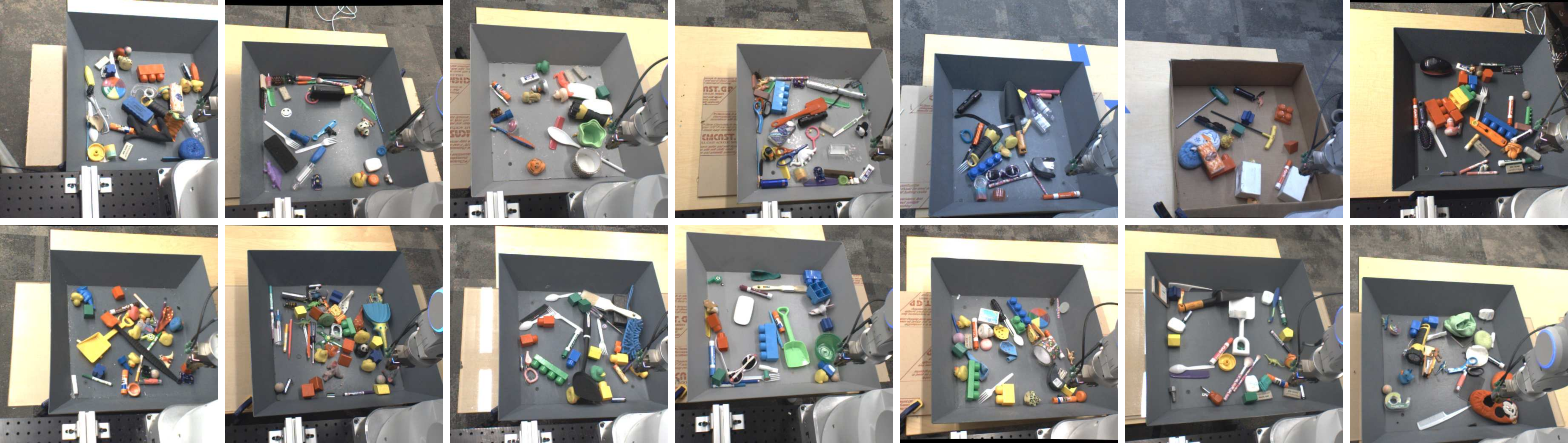}
        \end{center}
        \caption{Images from the cameras of each of the robots during training, with each robot holding the same joint configuration. Note the variation in the bin location, the difference in lighting conditions, the difference in pose of the camera relative to the robot, and the variety of training objects. \label{fig:kcam_images}}
	\vspace{-0.1in}
\end{figure*}

\begin{figure}
	\begin{center}
	\includegraphics[width=0.9\columnwidth]{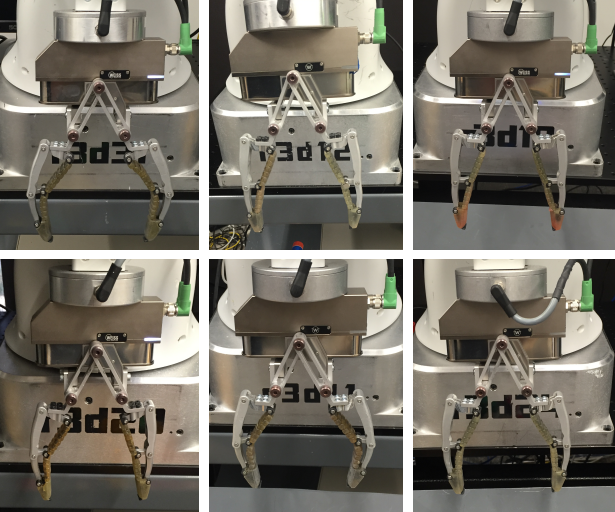}
	\end{center}
	\vspace{-0.1in}
	\caption{The grippers of the robots used for data collection at the end of our experiments. Different robots experienced different degrees of wear and tear, resulting in significant variation in gripper appearance and geometry. \label{fig:grippers}}
	\vspace{-0.1in}
\end{figure}

Our robotic manipulator platform consists of a lightweight 7 degree of freedom arm, a compliant, underactuated, two-finger gripper, and a camera mounted behind the arm looking over the shoulder. An illustration of a single robot is shown in Figure~\ref{fig:singlerobot}. The underactuated gripper provides some degree of compliance for oddly shaped objects, at the cost of producing a loose grip that is prone to slipping. An interesting property of this gripper was uneven wear and tear over the course of data collection, which lasted several months. Images of the grippers of various robots are shown in Figure~\ref{fig:grippers}, illustrating the range of variation in gripper wear and geometry. Furthermore, the cameras were mounted at slightly varying angles, providing a different viewpoint for each robot. The views from the cameras of all 14 robots during data collection are shown in Figure~\ref{fig:kcam_images}.

\subsection{Data Collection}
\label{sec:objects}

We collected about 800,000 grasp attempts over the course of two months, using between 6 and 14 robots at any given point in time, without any manual annotation or supervision. The only human intervention into the data collection process was to replace the object in the bins in front of the robots and turn on the system. The data collection process started with random motor command selection and $T = 2$.\footnote{The last command is always $\grasp_T = \emptyset$ and corresponds to closing the gripper without moving.} When executing completely random motor commands, the robots were successful on 10\% - 30\% of the grasp attempts, depending on the particular objects in front of them. About half of the dataset was collected using random grasps, and the rest used the latest network fitted to all of the data collected so far. Over the course of data collection, we updated the network 4 times, and increased the number of steps from $T = 2$ at the beginning to $T = 10$ at the end.

The objects for grasping were chosen among common household and office items, and ranged from a $4$ to $20$ cm in length along the longest axis. Some of these objects are shown in Figure~\ref{fig:kcam_images}. The objects were placed in front of the robots into metal bins with sloped sides to prevent the objects from becoming wedged into corners. The objects were periodically swapped out to increase the diversity of the training data.

Grasp success was evaluated using two methods: first, we marked a grasp as successful if the position reading on the gripper was greater than 1 cm, indicating that the fingers had not closed fully. However, this method often missed thin objects, and we also included a drop test, where the robot picked up the object, recorded an image of the bin, and then dropped any object that was in the gripper. By comparing the image before and after the drop, we could determine whether any object had been picked up.

\section{Experiments}
\label{sec:experiments}

\begin{figure}
	\includegraphics[width=0.99\columnwidth]{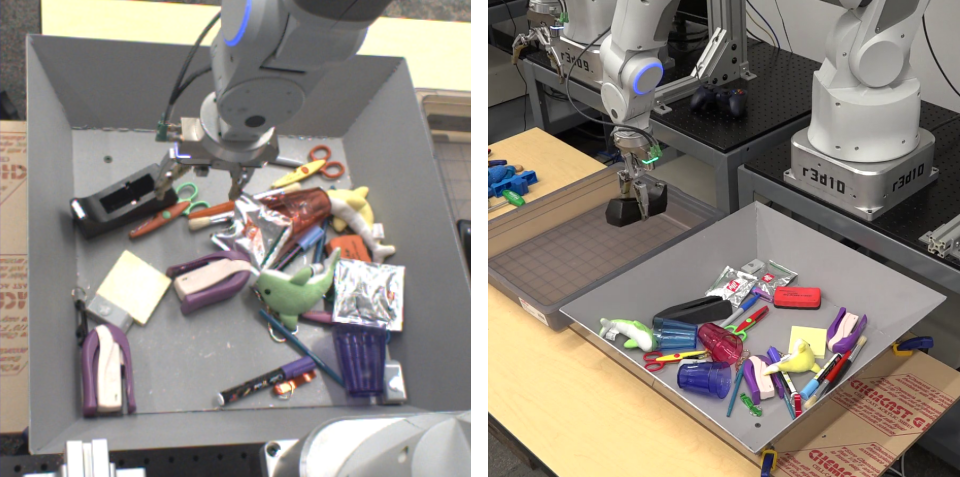}
	\caption{Previously unseen objects used for testing (left) and the setup for grasping without replacement (right). The test set included heavy, light, flat, large, small, rigid, soft, and translucent objects. \label{fig:test_objects}}
	\vspace{-0.1in}
\end{figure}

To evaluate our continuous grasping system, we conducted a series of quantitative experiments with novel objects that were not seen during training. The particular objects used in our evaluation are shown in Figure~\ref{fig:test_objects}. This set of objects presents a challenging cross section of common office and household items, including objects that are heavy, such as staplers and tape dispensers, objects that are flat, such as post-it notes, as well as objects that are small, large, rigid, soft, and translucent.

\subsection{Experimental Setup}

The goal of our evaluation was to answer the following questions: (1) does continuous servoing significantly improve grasping accuracy and success rate? (2) how well does our learning-based system perform when compared to alternative approaches? To answer question (1), we compared our approach to an open-loop method that observes the scene prior to the grasp, extracts image patches, chooses the patch with the highest probability of a successful grasp, and then uses a known camera calibration to move the gripper to that location. This method is analogous to the approach proposed by \citet{lg-sss-15}, but uses the same network architecture as our method and the same training set. We refer to this approach as ``open loop,'' since it does not make use of continuous visual feedback. To answer question (2), we also compared our approach to a random baseline method, as well as a hand-engineered grasping system that uses depth images and heuristic positioning of the fingers. This hand-engineered system is described in Appendix~\ref{app:hand}. Note that our method requires fewer assumptions than either of the two alternative methods: unlike \citet{lg-sss-15}, we do not require knowledge of the camera to hand calibration, and unlike the hand-engineered system, we do not require either the calibration or depth images.

We evaluated the methods using two experimental protocols. In the first protocol, the objects were placed into a bin in front of the robot, and it was allowed to grasp objects for 100 attempts, placing any grasped object back into the bin after each attempt. Grasping with replacement tests the ability of the system to pick up objects in cluttered settings, but it also allows the robot to repeatedly pick up easy objects. To address this shortcoming of the replacement condition, we also tested each system without replacement, as shown in Figure~\ref{fig:test_objects}, by having it remove objects from a bin. For this condition, which we refer to as ``without replacement,'' we repeated each experiment 4 times, and we report success rates on the first 10, 20, and 30 grasp attempts.


\begin{table}
	\begin{tabular}{| l | l | l | l |}
		\hline
		\parbox[c][0.4in][c]{0.2\columnwidth}{without\\ replacement} & \parbox[c][0.4in][c]{0.2\columnwidth}{first 10\\ ($N = 40$)}\!\! & \parbox[c][0.4in][c]{0.2\columnwidth}{first 20\\ ($N = 80$)}\!\! & \parbox[c][0.4in][c]{0.2\columnwidth}{first 30\\ ($N = 120$)}\!\! \\
		\hline
		random & 67.5\% & 70.0\% & 72.5\% \\
		hand-designed\!\! & 32.5\% & 35.0\% & 50.8\% \\
		open loop & 27.5\% & 38.7\% & 33.7\% \\
		our method & {\bf 10.0}\% & {\bf 17.5}\% & {\bf 17.5}\% \\
		\hline
		\hline
		\parbox[c][0.4in][c]{0.2\columnwidth}{with\\ replacement} & \multicolumn{3}{l |}{failure rate ($N = 100$)} \\
		\hline
		random & \multicolumn{3}{l |}{69\%} \\
		hand-designed\!\! & \multicolumn{3}{l |}{35\%} \\
		open loop & \multicolumn{3}{l |}{43\%} \\
		our method & \multicolumn{3}{l |}{{\bf 20\%}} \\
		\hline
	\end{tabular}
	\caption{Failure rates of each method for each evaluation condition. When evaluating without replacement, we report the failure rate on the first 10, 20, and 30 grasp attempts, averaged over 4 repetitions of the experiment. \label{tbl:results}}
\vspace{-0.1in}
\end{table}

\subsection{Comparisons}

The results are presented in Table~\ref{tbl:results}. The success rate of our continuous servoing method exceeded the baseline and prior methods in all cases. For the evaluation without replacement, our method cleared the bin completely after 30 grasps on one of the 4 attempts, and had only one object left in the other 3 attempts (which was picked up on the $31^\text{st}$ grasp attempt in 2 of the three cases, thus clearing the bin). The hand-engineered baseline struggled to accurately resolve graspable objects in clutter, since the camera was positioned about a meter away from the table, and its performance also dropped in the non-replacement case as the bin was emptied, leaving only small, flat objects that could not be resolved by the depth camera. Many practical grasping systems use a wrist-mounted camera to address this issue \cite{leeper2014}. In contrast, our approach did not require any special hardware modifications. The open-loop baseline was also substantially less successful. Although it benefited from the large dataset collected by our parallelized data collection setup, which was more than an order of magnitude larger than in prior work \cite{lg-sss-15}, it was unable to react to perturbations, movement of objects, and variability in actuation and gripper shape.\footnote{The absolute performance of the open-loop method is lower than reported by \citet{lg-sss-15}. This can be attributed to differences in the setup: different objects, grippers, and clutter.}

\begin{figure}
	\includegraphics[width=0.99\columnwidth]{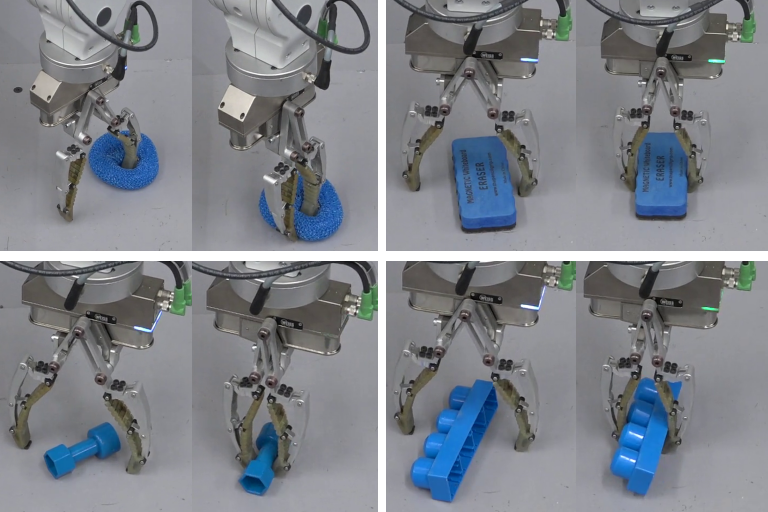}
	\caption{Grasps chosen for objects with similar appearance but different material properties. Note that the soft sponge was grasped with a very different strategy from the hard objects. \label{fig:hard_soft}}
	\vspace{-0.1in}
\end{figure}

\subsection{Evaluating Data Requirements}

\begin{table}
	\begin{tabular}{| l | l | l | l |}
		\hline
		\parbox[c][0.4in][c]{0.2\columnwidth}{without\\ replacement} & \parbox[c][0.4in][c]{0.15\columnwidth}{first 10\\ $N \!=\! 40$}\!\! & \parbox[c][0.4in][c]{0.15\columnwidth}{first 20\\ $N \!=\! 80$}\!\! & \parbox[c][0.4in][c]{0.15\columnwidth}{first 30\\ $N \!=\! 120$}\!\! \\
		\hline
		$12\%$: $M = $ 182,249 & 52.5\% & 45.0\% & 47.5\% \\
		$25\%$: $M = $ 407,729 & 30.0\% & 32.5\% & 36.7\% \\
		$50\%$: $M = $ 900,162 & 25.0\% & 22.5\% & 25.0\% \\
		$100\%$: $M = $ 2,898,410\!\! & {\bf 10.0}\% & {\bf 17.5}\% & {\bf 17.5}\% \\
		\hline
	\end{tabular}
	\caption{Failure rates of our method for varying dataset sizes, where $M$ specifies the number of images in the training set, and the datasets correspond roughly to the first eighth, quarter, and half of the full dataset used by our method. Note that performance continues to improve as the amount of data increases. \label{tbl:data}}
\vspace{-0.1in}
\end{table}

In Table~\ref{tbl:data}, we evaluate the performance of our model under the no replacement condition with varying amounts of data. We trained grasp prediction models using roughly the first $12\%$, $25\%$, and $50\%$ of the grasp attempts in our dataset, to simulate the effective performance of the model one eighth, one quarter, and one half of the way through the data collection process. Table~\ref{tbl:data} shows the size of each dataset in terms of the number of images. Note that the length of the trajectories changed over the course of data collection, increasing from $T=2$ at the beginning to $T=10$ at the end, so that the later datasets are substantially larger in terms of the total number of images. Furthermore, the success rate in the later grasp attempts was substantially higher, increasing from $10$ to $20\%$ in the beginning to around $70\%$ at the end (using $\epsilon$-greedy exploration with $\epsilon=0.1$, meaning that one in ten decisions was taken at random). Nonetheless, these results can be informative for understanding the data requirements of the grasping task. First, the results suggest that the grasp success rate continued to improve as more data was accumulated, and a high success rate (exceeding the open-loop and hand-engineered baselines) was not observed until at least halfway through the data collection process. The results also suggest that collecting additional data could further improve the accuracy of the grasping system, and we plan to experiment with larger datasets in the future.

\subsection{Qualitative Results}

Qualitatively, our method exhibited some interesting behaviors. Figure~\ref{fig:hard_soft} shows the grasps that were chosen for soft and hard objects. Our system preferred to grasp softer objects by embedding the finger into the center of the object, while harder objects were grasped by placing the fingers on either side. Our method was also able to grasp a variety of challenging objects, some of which are shown in Figure~\ref{fig:hard_objects}. Other interesting grasp strategies, corrections, and mistakes can be seen in our supplementary video: \url{https://youtu.be/cXaic_k80uM}

\begin{figure}
	\includegraphics[width=0.99\columnwidth]{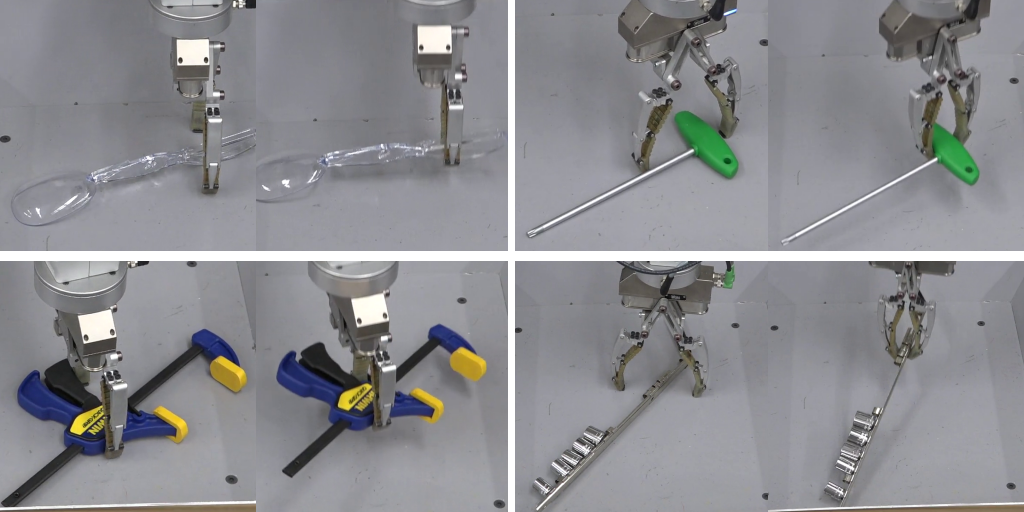}
	\caption{Examples of difficult objects grasped by our algorithm, including objects that are translucent, awkardly shaped, and heavy. \label{fig:hard_objects}}
	\vspace{-0.1in}
\end{figure}

\section{Discussion and Future Work}
\label{sec:discussion}

We presented a method for learning hand-eye coordination for robotic grasping, using deep learning to build a grasp success prediction network, and a continuous servoing mechanism to use this network to continuously control a robotic manipulator. By training on over 800,000 grasp attempts from 14 distinct robotic manipulators with variation in camera pose, we can achieve invariance to camera calibration and small variations in the hardware. Unlike most grasping and visual servoing methods, our approach does not require calibration of the camera to the robot, instead using continuous feedback to correct any errors resulting from discrepancies in calibration. Our experimental results demonstrate that our method can effectively grasp a wide range of different objects, including novel objects not seen during training. Our results also show that our method can use continuous feedback to correct mistakes and reposition the gripper in response to perturbation and movement of objects in the scene.

As with all learning-based methods, our approach assumes that the data distribution during training resembles the distribution at test-time. While this assumption is reasonable for a large and diverse training set, such as the one used in this work, structural regularities during data collection can limit generalization at test time. For example, although our method exhibits some robustness to small variations in gripper shape, it would not readily generalize to new robotic platforms that differ substantially from those used during training. Furthermore, since all of our training grasp attempts were executed on flat surfaces, the proposed method is unlikely to generalize well to grasping on shelves, narrow cubbies, or other drastically different settings. These issues can be mitigated by increasing the diversity of the training setup, which we plan to explore as future work.

One of the most exciting aspects of the proposed grasping method is the ability of the learning algorithm to discover unconventional and non-obvious grasping strategies. We observed, for example, that the system tended to adopt a different approach for grasping soft objects, as opposed to hard ones. For hard objects, the fingers must be placed on either side of the object for a successful grasp. However, soft objects can be grasped simply by pinching into the object, which is most easily accomplished by placing one finger into the middle, and the other to the side. We observed this strategy for objects such as paper tissues and sponges. In future work, we plan to further explore the relationship between our self-supervised continuous grasping approach and reinforcement learning, in order to allow the methods to learn a wider variety of grasp strategies from large datasets of robotic experience.

At a more general level, our work explores the implications of large-scale data collection across multiple robotic platforms, demonstrating the value of this type of automatic large dataset construction for real-world robotic tasks. Although all of the robots in our experiments were located in a controlled laboratory environment, in the long term, this class of methods is particularly compelling for robotic systems that are deployed in the real world, and therefore are naturally exposed to a wide variety of environments, objects, lighting conditions, and wear and tear. For self-supervised tasks such as grasping, data collected and shared by robots in the real world would be the most representative of test-time inputs, and would therefore be the best possible training data for improving the real-world performance of the system. So a particularly exciting avenue for future work is to explore how our method would need to change to apply it to large-scale data collection across a large number of deployed robots engaged in real world tasks, including grasping and other manipulation skills.

\subsection*{Acknowledgements}

We would like to thank Kurt Konolige and Mrinal Kalakrishnan for additional engineering and insightful discussions, Jed Hewitt, Don Jordan, and Aaron Weiss for help with maintaining the robots, Max Bajracharya and Nicolas Hudson for providing us with a baseline perception pipeline, and Vincent Vanhoucke and Jeff Dean for support and organization.

\bibliography{references}
\bibliographystyle{icml2016}

\appendix

\section{Servoing Implementation Details}
\label{app:servo}

In this appendix, we discuss the details of the inference procedure we use to infer the motor command $\grasp_t$ with the highest probability of success, as well as additional details of the servoing mechanism.

In our implementation, we performed inference using three iterations of cross-entropy method (CEM). Each iteration of CEM consists of sampling $64$ sample grasp directions $\grasp_t$ from a Gaussian distribution with mean $\mu$ and covariance $\Sigma$, selecting the $6$ best grasp directions (i.e. the $90^\text{th}$ percentile), and refitting $\mu$ and $\Sigma$ to these $6$ best grasps. The first iteration samples from a zero-mean Gaussian centered on the current pose of the gripper. All samples are constrained (via rejection sampling) to keep the final pose of the gripper within the workspace, and to avoid rotations of more than $180^\circ$ about the vertical axis. In general, these constraints could be used to control where in the scene the robot attempts to grasp, for example to impose user constraints and command grasps at particular locations.

Since the CNN $\prednet(\img_t,\grasp_t)$ was trained to predict the success of grasps on sequences that always terminated with the gripper on the table surface, we project all grasp directions $\grasp_t$ to the table height (which we assume is known) before passing them into the network, although the actual grasp direction that is executed may move the gripper above the table, as shown in Algorithm~\ref{alg:servo}. When the servoing algorithm commands a gripper motion above the table, we choose the height uniformly at random between $4$ and $10$ cm.

In our prototype, the dimensions and position of the workspace were set manually, by moving the arm into each corner of the workspace and setting the corner coordinates. In practice, the height of the table and the spatial extents of the workspace could be obtained automatically, for example by moving the arm until contact, or the user or higher-level planning mechanism.

\section{Determining Grasp Success}
\label{app:success}

We employ two mechanisms to determine whether a grasp attempt was successful. First, we check the state of the gripper after the grasp attempt to determine whether the fingers closed completely. This simple test is effective at detecting large objects, but can miss small or thin objects. To supplement this success detector, we also use an image subtraction test, where we record an image of the scene after the grasp attempt (with the arm lifted above the workspace and out of view), and another image after attempting to drop the grasped object into the bin. If no object was grasped, these two images are usually identical. If an object was picked up, the two images will be different.

\section{Details of Hand-Engineered Grasping System Baseline}
\label{app:hand}

The hand-engineered grasping system baseline results reported in Table~\ref{tbl:results} were obtained using perception pipeline that made use of the depth sensor instead of the monocular camera, and required extrinsic calibration of the camera with respect to the base of the arm. The grasp configurations were computed as follows: First, the point clouds obtained from the depth sensor were accumulated into a voxel map. Second, the voxel map was turned into a 3D graph and segmented using standard graph based segmentation; individual clusters were then further segmented from top to bottom into ``graspable objects'' based on the width and height of the region. Finally, a best grasp was computed that aligns the fingers centrally along the longer edges of the bounding box that represents the object. This grasp configuration was then used as the target pose for a task-space controller, which was identical to the controller used for the open-loop baseline.

\end{document}